\pdfoutput=1

\documentclass[11pt]{article}

\usepackage[]{acl}

\usepackage{times}
\usepackage{latexsym}

\usepackage[T1]{fontenc}

\usepackage[utf8]{inputenc}

\usepackage{microtype}

\usepackage{inconsolata}

\usepackage{graphicx}
\usepackage{tikz}
\usetikzlibrary{shapes,arrows}
\usetikzlibrary{positioning}
\usepackage{booktabs}
\usepackage{tabularx} 
\usepackage{makecell} 
\usepackage{natbib}

\usepackage{latexsym}
\usepackage{adjustbox}
\usepackage{colortbl}
\usepackage{xcolor}
\usepackage{todonotes}
\usepackage{mdframed}
\usepackage{afterpage}
\usepackage{multirow} 
\usepackage{tabulary}
\usepackage{array}
\usepackage{float}
\usepackage{xspace}
\usepackage{textcomp}

\newcommand*\rot{\rotatebox{90}}

\usepackage{paralist}
\usepackage{amsmath}

\newcommand{\ourmodel}{\textsc{CoSTA}\xspace}
\newcommand{\x}{\mathbf{x}\xspace}
\newcommand{\y}{\mathbf{y}\xspace}
\newcommand{\s}{\mathbf{s}\xspace}
\newcommand{\sx}{\mathbf{x}_{\mathbf{s}}\xspace}
\newcommand{\loss}{\mathcal{L}\xspace}

\title{\ourmodel: Code-Switched Speech Translation \\ using Aligned Speech-Text Interleaving}

\author{Bhavani Shankar, Preethi Jyothi, Pushpak Bhattacharyya \\
Indian Institute of Technology Bombay, India \\
\texttt{\{bhavanishankar, pjyothi, pb\}@cse.iitb.ac.in}}

\begin{document}
\maketitle
\begin{abstract}

Code-switching is a widely prevalent linguistic phenomenon in multilingual societies like India. Building speech-to-text models for code-switched speech is challenging due to limited availability of datasets. In this work, we focus on the problem of spoken translation (ST) of code-switched speech in Indian languages to English text. We present a new end-to-end model architecture \ourmodel that scaffolds on pretrained automatic speech recognition (ASR) and machine translation (MT) modules (that are more widely available for many languages). Speech and ASR text representations are fused using an aligned interleaving scheme and are fed further as input to a pretrained MT module; the whole pipeline is then trained end-to-end for spoken translation using synthetically created ST data. We also release a new evaluation benchmark for code-switched Bengali-English, Hindi-English, Marathi-English and Telugu-English speech to English text. \ourmodel significantly outperforms many competitive cascaded and end-to-end multimodal baselines by up to $3.5$ BLEU points.

\end{abstract}

\section{Introduction}
More than half of the world's population is presumed to be bilingual \cite{grosjean2021life}, often leading to code-switching (CS) in conversational speech, where speakers interweave words and phrases from multiple languages within a single utterance. 
Recent work has explored code-switching in automatic speech recognition (ASR) and machine translation (MT) fairly extensively. In contrast, spoken translation (ST) of code-switched speech has been somewhat under-explored. This is largely due to the lack of evaluation benchmarks for ST using code-switched speech and the predominantly monolingual bias of current state-of-the-art ASR/MT systems.


In this work, we present a new and effective end-to-end solution for ST of code-switched speech \ourmodel, starting from pretrained ASR and MT backbones. Simply cascading ASR and MT modules in a sequence is not very effective, since code-switched speech results in many ASR errors which cascade further via the MT module into the final ST predictions. We propose using a speech encoder for the input speech and a text encoder for the ASR text to derive speech and text representations, respectively. These sequences are force-aligned and the speech-text representations are interleaved according to the alignment. This merged representation is then fed as input to a pretrained MT module, and the entire pipeline is trained end-to-end with an ST objective.%
\footnote{At test time, we first produce an ASR output from the speech encoder which is subsequently used to create the interleaved speech-text embeddings.}
Aligned interleaving of speech-text representations is an important design choice in \ourmodel that is critical to deriving superior ST performance. 

Apart from a new ST model for code-switching, we release a new suite of code-switched evaluation sets for ST in Bengali-English, Hindi-English, Marathi-English and Telugu-English starting with ASR data from the Indicvoices dataset \cite{javed2024indicvoices}. We also release two new podcast-based evaluation sets for ST with more complex code-switching in Hindi-English and Telugu-English. (We also develop two new monolingual evaluation sets for ST in Telugu and Hindi, to evaluate how our model fares on largely monolingual inputs.) To the best of our knowledge, we are the first to release ST resources to translate from the four Indic code-switched language pairs to English. 

We train our model on a modest 30 hours of synthetic ST data, where translations are generated automatically from ASR ground-truth transcriptions. We compare our trained model against state-of-the-art cascaded baselines and end-to-end baselines such as SeamlessM4T \cite{DBLP:journals/corr/abs-2308-11596} and Whisper-ST \cite{DBLP:conf/icml/RadfordKXBMS23}, which have been trained on substantially larger datasets. Despite \ourmodel being trained in a low-resource setting with synthetically generated translations, our model achieves the best BLEU scores significantly outperforming the best end-to-end baseline by at most 3.5 points. Interestingly, our model also demonstrates robustness to code-switching, showing no significant degradation in performance even with increasing amounts of code-switching in a sentence.

In summary, our main contributions are:
\begin{enumerate}

    \item We propose \ourmodel, a new end-to-end architecture for code-switched ST. We introduce an interleaving technique that aligns speech and text embeddings, boosting translation accuracy for code-switched speech (detailed in Section~\ref{sec:ourmodel}).
    \item We release new ST evaluation benchmarks for four different languages code-switched with English: Marathi, Telugu, Bengali, and Hindi (detailed in Section~\ref{sec:datasets}).  
    \item We show many detailed ablation experiments and demonstrate how \ourmodel is robust to varying degrees of code-switching (detailed in Sections~\ref{sec:results} and~\ref{sec:ablations}). 
\end{enumerate}
Our code and datasets will be publicly released upon publication. We intend to release all the evaluation sets under the CC-BY-4.0 license\footnote{\url{https://creativecommons.org/licenses/by/4.0/}}.




\section{Related Work}

To overcome the limitations of traditional cascaded spoken translation (ST) systems, recent work has shifted towards end-to-end (E2E) architectures that directly translate speech into text in a different language, eliminating the need for intermediate transcription \cite{DBLP:journals/corr/BerardPSB16, DBLP:conf/interspeech/WeissCJWC17,DBLP:conf/interspeech/KanoS017, DBLP:conf/icassp/BerardBKP18, inaguma-etal-2020-espnet, wang-etal-2020-curriculum, zhao-etal-2021-mutual}. 
However, training such E2E models posed challenges due to the need for cross-modal, cross-lingual capabilities and the scarcity of labeled ST data compared to machine translation (MT) and automatic speech recognition (ASR).

To address these challenges, prior work extensively explored techniques to leverage small amounts of labeled ST data including pretraining \cite{DBLP:conf/interspeech/WeissCJWC17, DBLP:conf/icassp/BerardBKP18, bansal-etal-2019-pre, wang-etal-2020-curriculum, alinejad-sarkar-2020-effectively, DBLP:conf/aaai/DongYW0X0021, zhang-2021-zjus, tang-etal-2022-unified, DBLP:conf/icml/LeGW0LS23, DBLP:journals/corr/abs-2402-19333}, data augmentation \cite{park19e_interspeech, Gangi2019DataAF, DBLP:conf/iwslt/ShanbhogueXSZG23}, self-training \cite{pino20_interspeech, DBLP:conf/interspeech/WangWPBAC21, fang-etal-2022-stemm}, and using self-supervised pre-trained audio representations \cite{nguyen20_interspeech, DBLP:conf/interspeech/WuWPG20, DBLP:conf/interspeech/WangWPBAC21, tang-etal-2022-unified}. Recognizing limitations in the single encoder architecture, prior work explored enhancements such as employing a second encoder to extract semantic information from speech or incorporating both acoustic and textual information into a stacked encoder\cite{DBLP:conf/aaai/DongWZX0021, DBLP:conf/aaai/DongYW0X0021}. Multi-task frameworks have also been shown to enhance the robustness of ST models \cite{DBLP:conf/icassp/TangPWMG21, DBLP:conf/interspeech/YeW021, bhavsar-etal-2022-hmist, zhang-etal-2023-rethinking}.

To bridge the modality gap between speech and text, several methods like mutual learning \cite{zhao-etal-2021-mutual}, projection into a common representation space \cite{DBLP:conf/acl/HanWJL21, duquenne-etal-2022-modules}, modality matching \cite{DBLP:conf/interspeech/ChenZRRMBZ22}, contrastive learning \cite{DBLP:conf/naacl/YeWL22, DBLP:conf/emnlp/YinLZWTY23} and cross-modal regularization with scheduled sampling \cite{DBLP:conf/acl/FangF23a} have also been explored. In very recent work, multimodal models like SeamlessM4T~\cite{DBLP:journals/corr/abs-2308-11596}, Maestro \cite{DBLP:conf/interspeech/ChenZRRMBZ22}, mSLAM \cite{DBLP:journals/corr/abs-2202-01374} learn shared representations for speech and text and simultaneously support multiple speech-to-text tasks like ASR and ST.  

None of the above-mentioned prior works were focused on code-switched ST. \citet{weller-etal-2022-end} addressed this challenge by creating a code-switched corpus for Spanish-English and exploring both cascaded and end-to-end architectures for speech translation. \citet{DBLP:journals/corr/abs-2210-01512} proposed a unified Language Agnostic E2E ST model (LAST) that is well-suited for code-switched ST.  

\ourmodel distinguishes itself from prior work by bootstrapping on pretrained MT and ASR models and combining speech and text modalities using a new interleaving technique. We also release new ST evaluation sets for four language pairs, Telugu-English, Marathi-English, Hindi-English and Bengali-English, that have not been previously addressed in any prior work.





\section{\ourmodel: Model Architecture}
\label{sec:ourmodel}

To train \ourmodel, we assume access to a spoken translation (ST) corpus $\mathcal{D} = \{(\s^i,\x^i,\y^i)\}_{i=1}^N$ with $N$ training triples, where each triple consists of speech in a source language ($\s^i$), its transcript ($\x^i$) and its translation in a target language ($\y^i$). Since we do not have access to any ST training data for our code-switched languages of interest, we create $\mathcal{D}$ by starting from an ASR corpus of code-switched speech such as IndicVoices and synthetically generating English translations using a pretrained model (such as IndicTrans \cite{gala2023indictrans}).
\begin{figure}[h]
\centering
\includegraphics[width=0.45\textwidth]{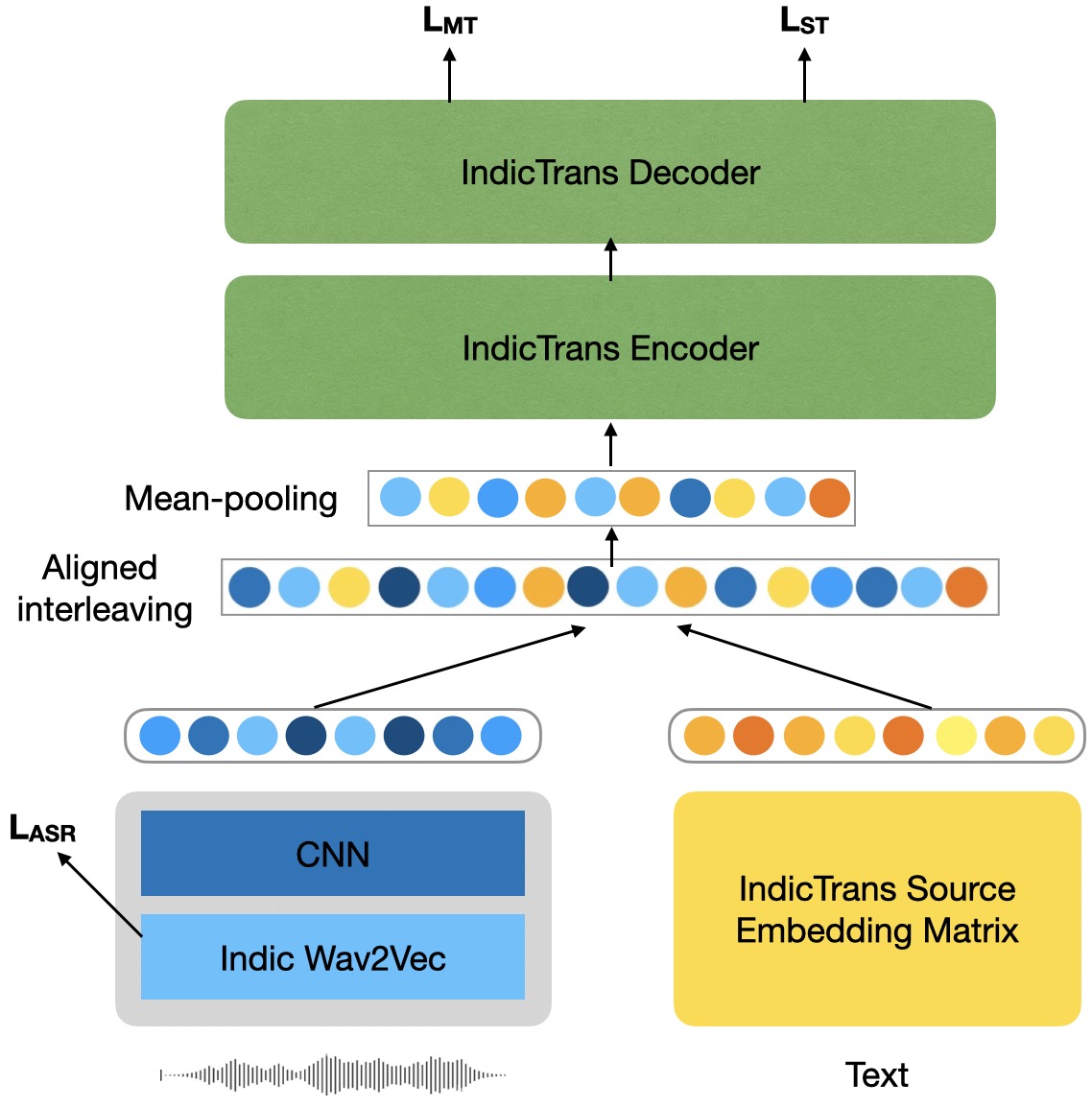}
\caption{Model with aligned interleaving, which aligns corresponding speech and text embeddings and interleaves them before passing them through the text encoder (IndicTrans encoder here).}
\label{fig:image1}
\end{figure}

Figure~\ref{fig:image1} shows a schematic diagram of \ourmodel. We assume access to pretrained ASR and MT models, which are encoder-only and encoder-decoder models, respectively. We use fine-tuned Indic Wav2Vec~\cite{javed2022towards} as our speech encoder and IndicTrans2~\cite{gala2023indictrans} as our pretrained MT model. \ourmodel uses the Indic Wav2Vec speech encoder to transform the input speech $\s$ into a sequence of speech representations $\{s_1,\ldots,s_T\}$ of length $T$. The source transcript $\x$ is tokenized and transformed into a sequence of text embeddings $\{x_1,\ldots,x_M\}$ of length $M$, using the tokenizer and text embedding layer of IndicTrans2's encoder. To bridge the length discrepancy between speech representations and token embeddings, two convolutional layers with a stride of 4 each are added after the ASR encoder (as in~\citet{DBLP:conf/naacl/YeWL22}). The resulting features are of dimensionality $d \times T/4$ where $d$ is the dimensionality of the encoder representations and the time dimension is reduced by a factor of 4.

\paragraph{Aligned interleaving.} Given an input speech representation sequence $\s = s_1,\ldots,s_T$ and its corresponding text token sequence $\x = x_1,\ldots,x_M$, how should we aggregate these representations to produce the target translation $\y$? We adopt the following simple strategy. A forced alignment \cite{pytorch_audio_forced_alignment} between $\s$ and $\x$ determines the number of speech frames aligned to each $x_j \in \x$. The representations of these aligned speech frames are averaged to compute $\bar{s}_j$. This gives us the following interleaved alignment: $\sx = \{(\bar{s}_1,x_1),\ldots,(\bar{s}_M,x_M)\}$. $\sx$ is fed as input to IndicTrans2's encoder and decoder modules. Simpler strategies like concatenating both sequences lead to degraded performance. Aligned interleaving of speech-text representations is critical to achieving high ST performance.

\paragraph{Training and Inference.} Our training loss is a combination of three objectives:
\[ \loss = \loss_{\text{ST}} + \lambda_1\loss_{\text{ASR}} + \lambda_2 \loss_{\text{MT}} \]

\begin{itemize}
    \item $\loss_{\text{ST}} = -\sum_{n=1}^{N} \log P(\y_n | \s_n, \x_n)$ is a cross entropy-based ST loss applied to the IndicTrans2 decoder. The IndicTrans2 encoder takes both $\s_n$ and $\x_n$ as inputs with aligned interleaving, and both the encoder and decoder are supervised using $\loss_{\text{ST}}$ to produce $\y_n$.
    \item $\loss_{\text{ASR}}$ is the standard CTC-based ASR loss~\cite{DBLP:conf/icml/GravesFGS06}. This is applied to the output of the encoder-only ASR model and encourages the model to perform well on the intermediate ASR task.
    \item $\loss_{\text{MT}} = -\sum_{n=1}^{N} \log P(\y_n | \x_n)$ is the standard cross-entropy MT loss to train the MT encoder and decoder, encouraging the model to perform well on the intermediate MT task.
\end{itemize}

The triplets in the training corpus $\mathcal{D}$ contain all the information needed to define all three of the above-mentioned loss terms. $\lambda_1$ and $\lambda_2$ are scaling factor hyperparameters for the loss terms that we tune on a validation set ($\lambda_1 = 1$ and $\lambda_2 = 1.5$).

During inference, the ASR transcript is derived from the ASR encoder-only model first, and subsequently force-aligned with the input speech to create the interleaved representation sequence.

\section{Dataset Details}
\label{sec:datasets}

\subsection{Training/Fine-tuning Data}

Due to the absence of existing speech-transcription-translation datasets for the code-switched languages Telugu-English, Hindi-English, Marathi-English, and Bengali-English, we sourced 30 hours of ASR data for each language from IndicVoices~\cite{javed2024indicvoices}. IndicVoices is an ASR dataset comprising 7348 hours of natural, spontaneous speech from 16237 speakers across 22 Indian languages, featuring monolingual and code-switched speech. We want to use monolingual data during training to show that we get performance gains on code-switched ASR without access to any code-switched speech during training. We translate the ground-truth transcripts of the 30-hour dataset into English using IndicTrans2 \cite{gala2023indictrans}, the current state-of-the-art MT model for Indic-to-English translation. We reiterate here that our training set consists of ground-truth ASR transcriptions and synthetically generated translations. 

\subsection{Code-switched Evaluation Sets}

To create code-switched evaluation sets, we extracted approx. two hours of speech-transcription data for each of Telugu, Hindi, Marathi, and Bengali from IndicVoices with fairly high Code-Mixing Index (CMI) scores. CMI metric quantifies the amount of code-switching in a corpus; CMI of 0 indicates monolingual inputs, and the maximum value of 0.5 indicates an equal mix of both matrix (e.g. Telugu) and embedded language (e.g. English) tokens. We translated these transcripts into English using IndicTrans2 and manually post-edited any errors. Table~\ref{tab:code-switched-eval-set} provides statistics for the four code-switched evaluation sets.


\begin{table}[h!]
\centering
\small
\addtolength{\tabcolsep}{-0.2em}
\begin{tabular}{lcccc}
\hline
\textbf{} & \textbf{Te} & \textbf{Hi} & \textbf{Mr} & \textbf{Bn} \\
\hline
\textbf{Duration (Hrs.)} & 2.3 & 2.5 & 2.2 & 2.1 \\
\textbf{Instances} & 587 & 728 & 575 & 624 \\
\textbf{\# Speakers} & 102 & 85 & 110 & 124 \\
\textbf{CMI Score} & 25.5\% & 22.1\% & 21.7\% & 23.3\% \\
\hline
\end{tabular}
\caption{Statistics of the code-switched evaluation set.}
\label{tab:code-switched-eval-set}
\end{table}

\paragraph{Podcast Evaluation Sets.} To further evaluate our models on a more challenging dataset, we obtained permission from Telugu\footnote{\href{https://rawtalks.in}{Telugu Podcast}} and Hindi podcasters\footnote{\href{https://www.pulybazi.in}{Hindi Podcast}} to create new code-switched evaluation sets, referred to as the ``podcast evaluation sets". This dataset is highly conversational, multi-speaker, code-switched and has many disfluencies, making it more challenging than the code-switched evaluation sets derived from IndicVoices. The CMI scores for both the sets exceed $30\%$, indicating a significantly higher amount of code-mixing. We manually annotated the transcripts for the podcast speech and created English translations after removing disfluencies from the corresponding transcripts. Statistics of the podcast evaluation sets are shown in Table~\ref{tab:podcast-eval-set}.

\begin{table}[h!]
\centering
\small
\addtolength{\tabcolsep}{-0.2em}
\begin{tabular}{lccc}
\hline
\textbf{Language} & \textbf{Duration (Hrs.)} & \textbf{Instances} & \textbf{CMI Score}\\
\hline
Telugu & 2.5 & 624 & 32.14\% \\
Hindi & 2.2 & 578 & 30.21\% \\
\hline
\end{tabular}
\caption{Statistics of the podcast evaluation set.}
\label{tab:podcast-eval-set}
\end{table}

\subsection{Monolingual Evaluation Sets}

We also extract monolingual evaluation sets from IndicVoices for each of Telugu and Hindi, and translated the speech transcripts into English using IndicTrans2. We created monolingual datasets as a control to check how well \ourmodel performs on them. The translations were manually verified to post-edit any errors in the machine-generated outputs. 
Statistics of the monolingual evaluation sets are presented in Table~\ref{tab:monolingual-eval-set}. Annotation guidelines for all these evaluation sets can be found in Appendix~\ref{sec:annot}. 
\begin{table}[h!]
\centering
\small
\addtolength{\tabcolsep}{-0.2em}
\begin{tabular}{lccc}
\hline
\textbf{Language} & \textbf{Duration (Hrs.)} & \textbf{Instances} & \textbf{\# Speakers} \\
\hline
Telugu & 2.5 & 581 & 126 \\
Hindi & 2.6 & 643 & 137 \\
\hline
\end{tabular}
\caption{Statistics of the monolingual evaluation set.}
\label{tab:monolingual-eval-set}
\end{table}

\section{Main Results}
\label{sec:results}
\begin{table*}[t!]
\footnotesize
\centering
\begin{tabular}{lp{4.8cm}p{0.8cm}p{0.8cm}p{0.8cm}p{0.8cm}p{0.8cm}p{0.8cm}p{0.8cm}p{0.8cm}}
\toprule
    & \textbf{Experiment} & \multicolumn{2}{c}{\textbf{Mr-En}} & \multicolumn{2}{c}{\textbf{Te-En}} & \multicolumn{2}{c}{\textbf{Bn-En}} & \multicolumn{2}{c}{\textbf{Hi-En}}\\
\cmidrule(lr){3-4} \cmidrule(lr){5-6} \cmidrule(lr){7-8} \cmidrule(lr){9-10}
    & & \textbf{BLEU} & \textbf{WER} & \textbf{BLEU} & \textbf{WER} & \textbf{BLEU} & \textbf{WER} & \textbf{BLEU} & \textbf{WER}\\
\midrule
\multirow{5}{*}{\textit{\rot{Cascaded}}}&IndicWav2Vec + NLLB & 9.21 & 43.12 & 11.23 & 37.95 & 13.24 & 38.90 & 17.82 & 33.21\\ 
&Seamless (ASR) + NLLB & 8.32 & 45.60 & 9.80 & 38.23 & 14.77 & 37.56 & 17.80 & 35.10\\ 
&Seamless (ASR + MT) & 9.29 & 45.60 & 10.12 & 38.23 & 15.16 & 37.56 & 18.12 & 35.10\\ 
&IndicWav2Vec + IndicTrans & 14.32 & 43.12 & 15.65 & 37.95 & 17.01 & 38.90 & 17.42 & 33.21\\ 
&IndicWav2Vec (ASR) + IndicTrans (MT) FT & 15.18 & 41.97 & 16.76 & 36.53 & 18.19 & 37.81 & 19.74 & 32.90\\ \hline
\multirow{7}{*}{\textit{\rot{End-to-End}}}&Whisper ST & 14.60 & 46.54 & 19.35 & 44.15 & 22.31 & 45.05 & 24.36 & 36.15\\ 
&Seamless E2E & 18.30 & 43.60 & 26.77 & 38.23 & 25.61 & 37.56 & 27.30 & 35.10\\ 
&Seamless E2E FT & 19.62 & 42.21 & 27.54 & 36.10 & 27.93 & 37.56 & 28.99 & 34.11\\ 
&Seamless FT MT+ST & 19.50 & 42.13 & 27.10 & 36.11 & 27.40 & 37.40 & 28.10 & 34.09\\ 
&Seamless FT ASR+ST & \textbf{19.66} & \textbf{40.75} & \textbf{27.83} & \textbf{35.61} & \textbf{28.65} & \textbf{35.15} & \textbf{29.65} & \textbf{32.20}\\ \cmidrule(lr){2-10}
& \ourmodel & \cellcolor{green!20}\textbf{21.43} & \cellcolor{green!20}\textbf{38.58} & \cellcolor{green!20}\textbf{29.87} & \cellcolor{green!20}\textbf{34.37} & \cellcolor{green!20}\textbf{31.05} & \textbf{34.89} & \cellcolor{green!20}\textbf{33.12} & \textbf{32.19}\\ 
\bottomrule
\end{tabular}
\caption{Comparison of cascaded and E2E baselines with \ourmodel for Marathi (Mr), Telugu (Te), Bengali (Bn), and Hindi (Hi) on the code-switched evaluation set. BLEU and WER scores are reported. The best baseline is in bold, with statistically significant improvements (at $p < 0.01$ using the Wilcoxon signed rank test) highlighted in green.}
\label{tab:cs_results}
\end{table*}
\paragraph{Cascaded Baselines.} A cascaded ST baseline consists of a state-of-the-art ASR followed by an MT system that translates the ASR transcript. For ASR, we used IndicWav2Vec \cite{javed2022towards}, a multilingual speech model pre-trained on 40 Indian languages and fine-tuned for ASR on 9 Indian languages. We also leveraged the ASR capabilities of SeamlessM4T-v2, a multimodal model that can take either speech or text as input for translation~\cite{DBLP:journals/corr/abs-2308-11596} (henceforth referred to as Seamless). For MT, we experimented with two state-of-the-art multilingual MT models, NLLB \cite{DBLP:journals/corr/abs-2207-04672} and IndicTrans2 \cite{gala2023indictrans} (henceforth referred to as IndicTrans). Additionally, we set up a Seamless (ASR + MT) cascaded baseline where Seamless is used for both ASR and MT. The above-mentioned baselines are all used zero-shot. We also fine-tuned the best cascaded baseline (IndicWav2vec + IndicTrans) on our 30-hour training dataset; IndicWav2vec is finetuned on the speech-transcription pairs for ASR and IndicTrans is fine-tuned on the transcription-translation pairs for MT. 
\begin{table}[b!]
\centering
\small
\begin{tabular}{lp{3.8cm}cc}
\toprule
 &   \textbf{Experiment} & \textbf{Te-En} & \textbf{Hi-En} \\
\midrule
\multirow{5}{*}{\textit{\rot{Cascaded}}}& IndicWav2Vec + NLLB & 11.82 & 12.31 \\ 
& Seamless (ASR) + NLLB & 08.91 & 11.56\\ 
& Seamless (ASR + MT) & 09.98 & 12.34\\ 
& IndicWav2Vec + IndicTrans & 14.97 & 15.20\\ 
& IndicWav2Vec (ASR) + IndicTrans (MT) FT  & 15.16 & 15.95\\ \hline 
\multirow{7}{*}{\textit{\rot{End-to-End}}}& Whisper ST & 17.68 & 21.08 \\ 
& Seamless E2E & 25.76 & 26.12 \\ 
& Seamless E2E FT & 26.49 & \textbf{27.01}\\ 
& Seamless FT MT+ST & 26.12 & 26.40 \\ 
& Seamless FT ASR+ST & \textbf{26.87} & 26.93 \\\cmidrule(lr){2-4}
& \ourmodel  & \cellcolor{green!20}\textbf{28.75} & \cellcolor{green!20}\textbf{29.46} \\ 
\bottomrule
\end{tabular}
\caption{Comparison of cascaded and E2E baselines with \ourmodel for the languages Telugu and Hindi on the podcast evaluation set. BLEU scores are reported. The best baseline is in bold, with statistically significant improvements (at $p < 0.01$ using the Wilcoxon signed rank test) highlighted in green.}
\label{tab:podcast_results}
\end{table}

\paragraph{End-to-End (E2E) Baselines.} For E2E baselines, we used two state-of-the-art E2E ST models, Whisper \cite{DBLP:conf/icml/RadfordKXBMS23} and Seamless \cite{DBLP:journals/corr/abs-2308-11596} that we refer to as ``Whisper ST" and ``Seamless E2E", respectively. While Seamless itself is a strong baseline, comparable in performance to GPT-4o \cite{openai2023hellogpt}, we enhanced it further by fine-tuning it on our 30 hr training set to establish stronger baselines. ``Seamless E2E" represents the use of SeamlessM4T-v2 without any additional fine-tuning. ``Seamless E2E FT" indicates that Seamless was fine-tuned directly for ST. ``Seamless FT MT+ST" refers to fine-tuning first for MT using only the transcription-translation pairs and subsequently fine-tuning on ST using speech-translation pairs from our 30-hr dataset. Finally, ``Seamless FT ASR+ST" refers to an initial fine-tuning for ASR (using speech-transcription pairs) followed by further fine-tuning for ST (using speech-translation pairs). 

\paragraph{Results.} Tables~\ref{tab:cs_results} and~\ref{tab:podcast_results} show the results of \ourmodel in comparison with the cascaded and E2E Baselines on the IndicVoices and podcast evaluation sets, respectively. We show both BLEU scores of the final translations, as well as word error rates (WERs) of the ASR transcripts. \ourmodel significantly outperforms strong E2E baselines wrt BLEU scores on all four language pairs in Table~\ref{tab:cs_results} and both podcast evaluation sets in Table~\ref{tab:podcast_results}. It is also evident that E2E baselines are significantly better than cascaded baselines for code-switched evaluation sets. From the WERs in Table~\ref{tab:cs_results}, we also observe that significant improvements in BLEU scores are not contingent on obtaining significant reductions in WER (e.g., Bn and Hi). 
\looseness=-1

Table~\ref{tab:ncs_results} shows the results of all systems on the two monolingual evaluation sets. Here, we find cascaded baselines to be significantly better than E2E baselines. Despite being E2E, \ourmodel is statistically comparable in performance to the best cascaded baseline for monolingual evaluation sets.

\begin{table}[h]
\small
\centering
\begin{tabular}{lp{3.8cm}cc}
\toprule
   &  \textbf{Experiment} & \textbf{Te-En} & \textbf{Hi-En}\\
\midrule
\multirow{5}{*}{\textit{\rot{Cascaded}}}& IndicWav2Vec + NLLB & 26.72 & 27.30 \\ 
&Seamless (ASR) + NLLB & 24.15 & 25.09 \\ 
&Seamless (ASR + MT) & 23.21 & 25.11 \\ 
&IndicWav2Vec + IndicTrans & 28.56 & 28.78 \\ 
& IndicWav2Vec (ASR) + IndicTrans (MT) FT  & \textbf{29.75} & \textbf{29.90} \\ \hline 
\multirow{7}{*}{\textit{\rot{End-to-End}}}&Whisper ST & 19.21 & 22.12 \\ 
&Seamless E2E & 24.45 & 27.54  \\ 
&Seamless E2E FT & 25.43 & 28.23 \\ 
&Seamless FT MT+ST & 25.50 & 28.70\\ 
&Seamless FT ASR+ST & 26.01 & 28.65 \\\cmidrule(lr){2-4}
& \ourmodel  & 29.16 & 29.43 \\ 
\bottomrule
\end{tabular}
\caption{Comparison of cascaded and E2E baselines with \ourmodel for the languages Telugu (Te) and Hindi (Hi), Marathi (Mr), and Bengali (Bn) on the monolingual evaluation set. We report BLEU scores. We see that cascaded models outperform E2E models when the input is not code switched.}
\label{tab:ncs_results}
\end{table}

\section{Ablations and Other Experiments}
\label{sec:ablations}

\subsection{Evaluation of Code-switched Span Accuracy}
We claim that the improvements in BLEU scores using \ourmodel are aided by improved translations of code-switched spans. To empirically verify this claim, we aim to identify what fraction of English spans in the ground-truth ASR transcriptions appear as-is in the predicted English translations. 
First, we isolate all English spans by comparing the ground-truth ASR transcriptions and reference translations. Let us call these reference spans. Given a predicted translation, we check how many English spans in it exactly match the reference spans and in order. (Figure~\ref{tab:cs_span_eval} in Appendix~\ref{sec:cs_span} further explains this calculation with an example.) If two English spans in a translation match two of four reference spans in the correct order, the match percentage will be $50\%$. We note here this is an exact match of the English words and does not account for correct synonyms or paraphrases, thus making it a stricter evaluation of accuracy.
Table~\ref{tab:cs_span} shows these English span accuracies using \ourmodel, the best E2E baseline and the best cascaded baseline. \ourmodel achieves the highest exact match among the three (highlighted in bold), thus indicating that it is most successful in accurately retaining English words from the ASR transcriptions in the predicted translations.


\begin{table}[h]
    \small
    \centering
    \setlength\tabcolsep{3pt} 
    \begin{tabular}{lccc}
        \toprule
        & \multicolumn{1}{c}{\textbf{Best Cascaded}} & \multicolumn{1}{c}{\textbf{Seamless}} & \multicolumn{1}{c}{\textbf{\ourmodel}} \\
        \midrule
        Te-En & 8.9\% & 54.3\%  & \textbf{57.9\%} \\[0.1cm]
        Hi-En & 14.1\% & 55.1\% & \textbf{59.6\%} \\[0.1cm]
        Mr-En & 7.5\% & 48.1\% & \textbf{49.7\%} \\[0.1cm]
        Bn-En & 8.4\% & 51.9\% & \textbf{54.2\%} \\[0.1cm]
        \bottomrule
    \end{tabular}
    \caption{Comparison of \ourmodel with the best cascaded model (IndicWav2Vec (ASR) + IndicTrans (MT) FT) and the best E2E model (Seamless FT ASR+ST). }
    \label{tab:cs_span}
\end{table}

\subsection{Robustness of \ourmodel's Performance to Amount of Code-switching}

\begin{table}[b!]
\centering
\small
\begin{tabular}{ccc}
\toprule
\textbf{Bin (English Words)} & \textbf{Te-En} & \textbf{Hi-En} \\  \midrule
3                            & 29.82                      & 32.94                     \\
5                            & 29.83                      & 33.28                     \\ 
7                            & 29.54                      & 33.27                     \\ 
10                           & 29.91                      & 33.04                     \\ \bottomrule
\end{tabular}
\caption{BLEU scores for the four bins with varying numbers of English words in Telugu and Hindi Speech.}
\label{tab:bins}
\end{table}

We assess the correlation between the number of English words in a sentence and the model’s score by using a linear model to determine $R^2$ values. Four distinct bins, each containing 50 sentences from the code-switched evaluation set are created, ensuring each sentence is at least 12 words long. The bins comprise sentences with 3, 5, 7, and 10 English words, respectively. Table~\ref{tab:bins} shows BLEU scores for the four bins for Telugu and Hindi. Our findings indicate nearly no correlation between the number of English words (indicating degree of code-switching) and the model’s score ($R^2 = 0.006$ for Telugu and $R^2 = 0.016$ for Hindi). This means that the models are fairly robust to the amount of code-switching in a sentence.

\subsection{Combining Speech and Text}

We compare different approaches for aggregating speech and text embeddings before passing them through the IndicTrans encoder. We evaluate four strategies: interleaving mean-pooled speech embeddings and text embeddings, starting with either speech or text, appending mean-pooled speech embeddings either before or after the text embeddings. We fine-tune our Telugu and Hindi Models with 30 hour data using all four approaches. Our findings in Table~\ref{tab:interleaving_vs_appending} indicate that interleaving consistently outperforms appending, with statistically significant improvements (at $p < 0.01$, Wilcoxon signed rank test). However, no discernible difference in performance was observed between starting the interleaving process with speech or text embeddings. 

\begin{table}[h]
    \small
    \centering
    \setlength\tabcolsep{3pt} 
    \begin{tabular}{lcc}
        \toprule
        & \multicolumn{1}{c}{\textbf{Te-En}} & \multicolumn{1}{c}{\textbf{Hi-En}} \\
        \midrule
        Append (Speech First) & 18.98 & 21.16 \\[0.1cm]
        Append (Text First) & 17.91 & 20.75 \\[0.1cm]
        Interleave (Speech First) & \textbf{29.87} &  \textbf{33.12} \\[0.1cm]
        Interleave (Text First) & \textbf{29.45} & \textbf{33.05} \\[0.1cm]
        \bottomrule
    \end{tabular}
    \caption{Comparison of different speech and text fusion strategies (two appending vs two interleaving).}
    \label{tab:interleaving_vs_appending}
\end{table}

\subsection{Using only $\mathcal{L}_{\text{ST}}$ loss compared to using all three losses ($\mathcal{L}_{\text{ST}}$, $\mathcal{L}_{\text{ASR}}$, $\mathcal{L}_{\text{MT}}$)}

Table~\ref{tab:onlySTloss} shows the results for all four languages on three different models fine-tuned on the 30 hour train set, using only the ST loss versus using all three losses (with $\lambda_1 = 1$ and $\lambda_2 = 1.5$). Even with using only the ST loss, our model shows significant improvement (at $p < 0.01$ using wilcoxon signed rank test) over Seamless, and using all the three losses ST, MT and ASR further significantly improves the BLEU scores.

\begin{table}[h]
    \small
    \centering
    \setlength\tabcolsep{3pt} 
    \begin{tabular}{lccc}
        \toprule
        & \multicolumn{1}{c}{\textbf{Seamless FT}} & \multicolumn{1}{c}{\textbf{Only ST Loss}} & \multicolumn{1}{c}{\textbf{\ourmodel}} \\
        \midrule
        Telugu & 27.54 & 28.96 & \textbf{29.87}\\[0.1cm]
        Hindi & 28.99 & 32.23 & \textbf{33.12} \\[0.1cm]
        Marathi & 19.62 &  20.71 & \textbf{21.43}\\[0.1cm]
        Bengali & 27.90 & 30.53 & 31.05\\[0.1cm]
        \bottomrule
    \end{tabular}
    \caption{We compare \ourmodel, Seamless E2E FT, and an ST loss-only model on four languages using the code-switched evaluation set. Results significantly better than both Seamless and ST loss models are in bold.
}
    \label{tab:onlySTloss}
\end{table}

\subsection{Size of Fine-tuning Dataset}

We gradually increased the size of the fine-tuning dataset from 5 hours to 30 hours to monitor BLEU scores on the code-switched evaluation sets as a function of size. Table~\ref{tab:ft_hours} shows that the BLEU scores stabilized for both \ourmodel and Seamless (finetuned) with about 30 hours of fine-tuning data. 
While Seamless outperformed \ourmodel with 5 and 10 hours of fine-tuning data, \ourmodel achieved statistically significant improvements (at $p < 0.01$, Wilcoxon signed rank test) over Seamless using data of size 15 hours (and more).

\begin{table}[h]
    \small
    \centering
    \setlength\tabcolsep{3pt} 
    \begin{tabular}{lcc|cc}
        \toprule
        \textbf{FT Data}& \multicolumn{2}{c|}{\textbf{Te-En}} & \multicolumn{2}{c}{\textbf{Hi-En}} \\
        \midrule
        & \textbf{\ourmodel} & \textbf{Seamless} &  \textbf{\ourmodel} & \textbf{Seamless} \\
        \midrule
        5 hrs & 23.90 & \textbf{26.85} & 25.12 & \textbf{27.81} \\[0.1cm]
        10 hrs & 27.16 & 27.10 & 27.35 &  \textbf{28.43}\\[0.1cm]
        15 hrs & \textbf{28.65} &  27.23 & \textbf{29.27} &  28.65\\[0.1cm]
        20 hrs & \textbf{29.55}    & 27.34 & \textbf{31.54}  & 28.76\\[0.1cm]
        25 hrs & \textbf{29.72}    & 27.52 & \textbf{32.91}  & 28.90\\[0.1cm]
        30 hrs & \textbf{29.87}  & 27.54 & \textbf{33.12}  & 28.99\\[0.1cm]
        \bottomrule
    \end{tabular}
    \caption{Comparison of the BLEU Scores with \ourmodel and seamless with varying numbers of hours of fine-tuning data.  We do this experiment on the languages Telugu and Hindi. }
    \label{tab:ft_hours}
\end{table}

\subsection{Mean-Pooling vs. Direct Interleaving}

We examine the impact of speech embedding aggregation on \ourmodel's performance. We compare `Mean-Pooling' where speech embeddings corresponding to a text embedding are averaged (mean-pooled) before being interleaved with the text embedding and passed to the IndicTrans Encoder with `Direct Interleaving' where speech embeddings are directly interleaved with the text embedding without mean-pooling. We conducted this comparison using varying amounts of fine-tuning data from 5 hours to 30 hours. Evaluations were performed on the Telugu IndicVoices code-switched evaluation set. Table~\ref{tab:meanpool} shows that mean-pooling speech embeddings consistently outperforms the direct interleaving approach, regardless of the amount of fine-tuning data used.

\begin{table}[h]
    \small
    \centering
    \setlength\tabcolsep{3pt} 
    \begin{tabular}{lcc}
        \toprule
       \textbf{FT Data} & \multicolumn{1}{c}{\textbf{Mean Pooling}} & \multicolumn{1}{c}{\textbf{Direct Interleaving}} \\
        \midrule
        5 hrs & \textbf{23.90} & 23.81 \\[0.1cm]
        10 hrs & \textbf{27.16} & 24.15 \\[0.1cm]
        15 hrs & \textbf{28.65} &  24.76 \\[0.1cm]
        20 hrs & \textbf{29.55} & 25.51 \\[0.1cm]
        25 hrs & \textbf{29.72} & 25.98 \\[0.1cm]
        30 hrs & \textbf{29.87} & 26.40 \\[0.1cm]
        \bottomrule
    \end{tabular}
    \caption{Comparison of BLEU scores on the Telugu code-switched evaluation set: Mean Pooling vs. Direct Interleaving of speech embeddings during aligned interleaving, using varying amounts of fine-tuning data.}
    \label{tab:meanpool}
\end{table}

\subsection{Cross-Domain Generalization Using Kathbath Data}

To assess the generalizability of \ourmodel when using a corpus different from IndicVoices during training, we experiment with using a fine-tuning set from Kathbath~\cite{DBLP:conf/aaai/JavedBRKKK23}. Kathbath comprises 1,684 hours of labeled read speech spanning 12 Indian languages. We trained Telugu and Hindi models using 30 hours of Kathbath data and translating ground-truth transcripts into English using IndicTrans2 to create speech-transcript-translate pairs. Table~\ref{tab:cs_results_kathbath} shows the comparison between \ourmodel, and all the cascaded and end-to-end baseline models when fine-tuned on Kathbath and evaluated on the code-switched test sets from IndicVoices. We observe that our model significantly outperforms all baselines, demonstrating its ability to generalize well when trained on cross-domain data.
\looseness=-1
\begin{table}[h!]
\centering
\small
\begin{tabular}{lp{3.8cm}cc}
\toprule
 &   \textbf{Experiment} & \textbf{Te-En} & \textbf{Hi-En} \\
\midrule
\multirow{5}{*}{\textit{\rot{Cascaded}}}& IndicWav2Vec + NLLB & 11.23 & 17.82\\ 
& Seamless (ASR) + NLLB & 9.80 & 17.80\\ 
& Seamless (ASR + MT) & 10.12 & 18.12\\ 
& IndicWav2Vec + IndicTrans & 15.65 & 17.42\\ 
& IndicWav2Vec (ASR) + IndicTrans (MT) FT  & 16.68 & 19.85\\ \hline 
\multirow{7}{*}{\textit{\rot{End-to-End}}}& Whisper ST & 19.35 & 24.36 \\ 
& Seamless E2E & 26.77 & 27.30 \\ 
& Seamless E2E FT & 27.38 & 29.05\\ 
& Seamless FT MT+ST & 27.23 & \textbf{29.21} \\ 
& Seamless FT ASR+ST & \textbf{28.05} & 29.13 \\\cmidrule(lr){2-4}

& \ourmodel  & \cellcolor{green!20}\textbf{28.54} & \cellcolor{green!20}\textbf{33.56} \\ 
\bottomrule
\end{tabular}
\caption{BLEU comparisons of all the baselines with \ourmodel for code-switched Telugu (Te-En) and Hindi (Hi-En) using Kathbath fine-tuning data. The best baseline is in bold, with statistically significant improvements (at $p < 0.01$ using the Wilcoxon signed rank test) highlighted in green.}
\label{tab:cs_results_kathbath}
\end{table}

\subsection{Projected Concatenation}

We compared our interleaving technique with an alternative where the mean-pooled speech embeddings (dimension $d$) and their corresponding text embeddings (dimension $d$) are concatenated, resulting in embeddings of dimension $2d$. These concatenated embeddings are then projected back to dimension $d$ using a single transformer encoder (with 16 attention heads). The resulting embeddings are then passed through the IndicTrans encoder. We refer to this technique as Projected Concatenation. Table~\ref{tab:concat+project} shows performance on all four code-switched evaluation sets after training using both interleaving strategies. While projected concatenation outperforms the Seamless E2E baseline, our interleaving module proves to be significantly better on all four evaluation sets.

\begin{table}[h]
    \small
    \centering
    \setlength\tabcolsep{3pt} 
    \begin{tabular}{lcc}
        \toprule
        & \multicolumn{1}{c}{\textbf{\ourmodel}} & \multicolumn{1}{c}{\textbf{Projected Concatenation}} \\
        \midrule
        Te-En & \textbf{29.87} & 26.94 \\[0.1cm]
        Hi-En & \textbf{33.12} & 32.21 \\[0.1cm]
        Mr-En & \textbf{21.43} &  20.11 \\[0.1cm]
        Bn-En & \textbf{31.05} & 29.81\\[0.1cm]
        \bottomrule
    \end{tabular}
    \caption{BLEU comparison of \ourmodel with Projected Concatenation that merge and project the speech-text embeddings using a learnable projection layer. We evaluate on four code-switched evaluation sets. Statistically significant improvements are highlighted in bold.}
    \label{tab:concat+project}
\end{table}

\subsection{Teacher Forcing vs Scheduled Sampling}

During training, we employ teacher forcing and pass the ground truth ASR transcripts through the text embedding module. However, during inference we rely on ASR transcripts generated by the ASR head of our model. To bridge this gap between training and inference, we compare teacher forcing with scheduled sampling that gradually introduces ASR-generated transcripts from our model into the text embedding layer by linearly decreasing the probability of using ground truth ($p$) at a fixed rate per epoch.
We train our model on 30 hours of training data for all four languages and evaluate on our code-switched evaluation sets. Our results in Table~\ref{tab:teacherforcing} demonstrate that teacher forcing significantly outperforms scheduled sampling for all languages.

\begin{table}[h]
    \small
    \centering
    \setlength\tabcolsep{3pt} 
    \begin{tabular}{lcc}
        \toprule
        & \multicolumn{1}{c}{\textbf{Teacher Forcing}} & \multicolumn{1}{c}{\textbf{Scheduled Sampling}} \\
        \midrule
        Te-En & \textbf{29.87} & 29.31 \\[0.1cm]
        Hi-En & \textbf{33.12} & 32.17 \\[0.1cm]
        Mr-En & \textbf{21.43} &  19.87 \\[0.1cm]
        Bn-En & \textbf{31.05} & 30.16\\[0.1cm]
        \bottomrule
    \end{tabular}
    \caption{Comparison of \ourmodel (teacher forcing) with the model trained using scheduled sampling. Higher BLEU for each language pair is highlighted in bold.}
    \label{tab:teacherforcing}
\end{table}

\section{Conclusion}
In this work, we propose a new technique \ourmodel for code-switched spoken translation where aligned speech and ASR text representations are fed as inputs to a pretrained MT model and finetuned end-to-end using ST data. We outperform multiple state-of-the-art cascaded and end-to-end baselines on code-switched evaluation sets in Telugu-English, Hindi-English, Marathi-English and Bengali-English. We also create a new evaluation benchmark for these language pairs for which no ST resources previously existed.  

\section*{Limitations}

\begin{enumerate}
    \item Our model necessitates labeled data comprising speech, transcription, and translation triplets for training. However, speech data is often scarce, particularly for low-resource languages making it challenging to acquire enough training data for our model.
    \item We assume fine-tuned ASR and MT models as the building blocks to our model.
    \item Our model still relies on the ASR module to transcribe speech into text during inference, which does not address the issue of high latency in the cascaded systems.
\end{enumerate}

\bibliography{custom}

\appendix

\section{Code-switched span accuracy Evaluation}
\label{sec:cs_span}

\begin{figure*}[h]
\centering
\includegraphics[width=\textwidth]{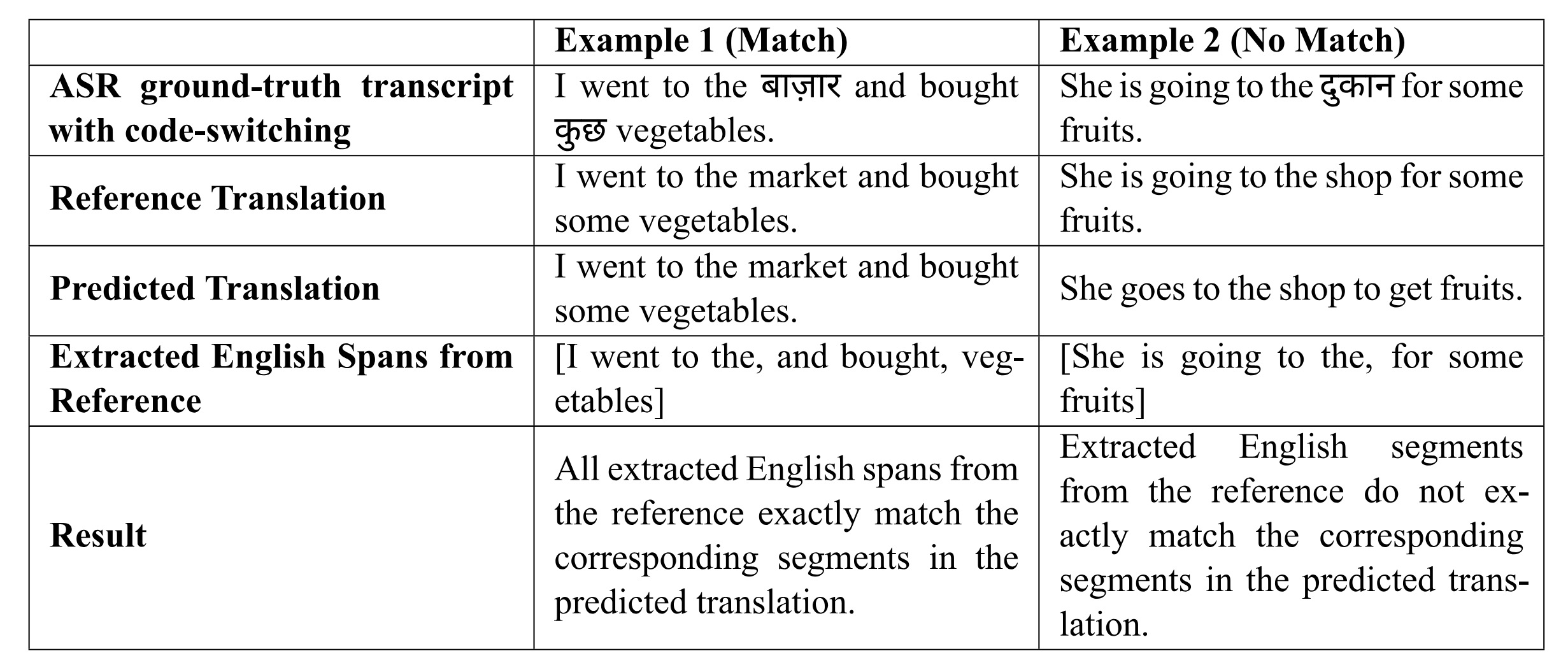}
\caption{To assess the accuracy of code-switched span translation, we evaluate the exact match between the English spans in the reference translation and the predicted translation. This involves identifying the English spans in the code-switched transcript and then comparing these spans with those in the predicted translations. It is important to note that this process is order-dependent.}
\label{tab:cs_span_eval}
\end{figure*}

Figure~\ref{tab:cs_span_eval} shows the process of calculating the exact match between spans of predicted and reference translations. We extract the spans from the code-switched ground-truth transcript. Corresponding spans are then matched, that means that the measure is order-dependant.

\section{Model output comparison}
\label{sec:output}

\begin{figure*}[h]
\centering
\includegraphics[width=\textwidth]{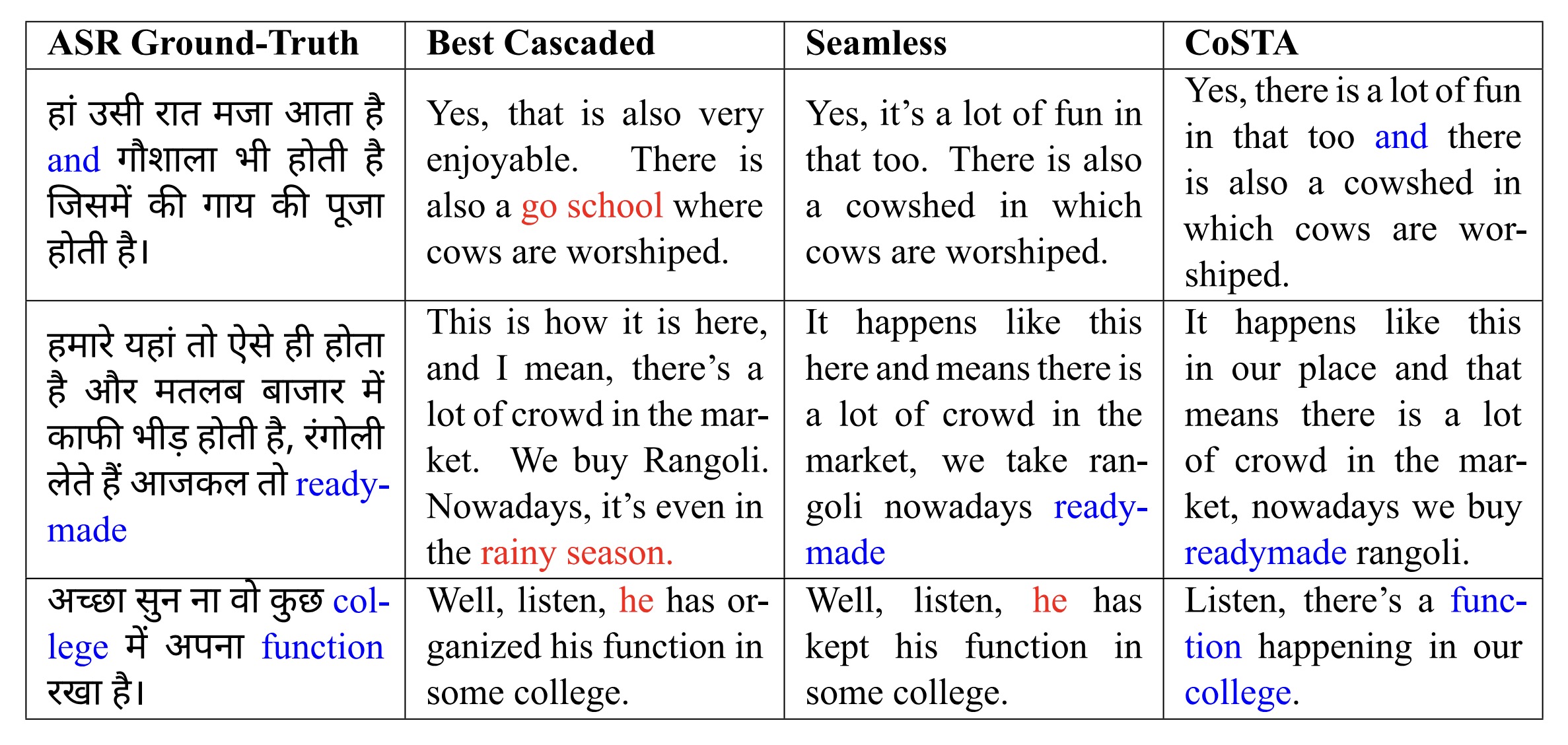}
\caption{Example generated outputs from the best hindi cascaded model (IndicWav2Vec for ASR combined with IndicTrans for MT, fine-tuned), the best seamless model (Seamless fine-tuned ASR+ST), and \ourmodel. Note that error propagation is observed in the cascaded model (highlighted in \textcolor{red}{red}), arising from multiple factors: an incorrect transcript in the first example, the English word \textit{ready-made} being incorrectly transcribed by the Hindi ASR model in the second example, and a machine translation error in the third example. Additionally, the English words uttered in the speech are correctly captured by \ourmodel (highlighted in \textcolor{blue}{blue}), unlike in the cascaded and seamless models.}
\label{fig:image2}
\end{figure*}

Figure~\ref{fig:image2} shows the outputs of three Hindi models: the best cascaded model (IndicWav2Vec for ASR combined with IndicTrans for MT, fine-tuned), the best seamless model (Seamless fine-tuned ASR+ST), and \ourmodel. We observe that the presence of English in Hindi speech introduces multiple propagation errors, resulting in erroneous English translations from the cascaded model, while the Seamless model and our end-to-end model attempt to mitigate this issue. In fact, \ourmodel outperforms the others in accurately capturing English words within Hindi speech.

\section{Lambda values for ASR and MT loss}
\label{sec:lambda}

To determine the values for $\lambda_1$ and $\lambda_2$, we conducted experiments using various combinations. We tested values of 0, 0.5, 1, and 1.5 for each parameter. We train a Telugu model with 30 hours of our fine-tuning data for each combination of $\lambda_1$ and $\lambda_2$, and then evaluated on the code-switched evaluation set. The highest score on the evaluation set was achieved when $\lambda_1 = 1$ and $\lambda_2 = 1.5$. The scores obtained with different values of $\lambda_1$ and $\lambda_2$ are presented in Table~\ref{tab:lambda}.

\begin{table}[h!]
    \small
    \centering
    \setlength\tabcolsep{3pt} 
    \begin{tabular}{llc}
        \toprule
         \multicolumn{1}{c}{\textbf{$\lambda_1$}} & \multicolumn{1}{c}{\textbf{\textbf{$\lambda_2$}}} & \multicolumn{1}{c}{\textbf{BLEU}} \\
        \midrule
        0&0&28.96\\[0.1cm]
        0.5&0&29.05\\[0.1cm]
        0&0.5&29.11\\[0.1cm]
        1&0&29.23\\[0.1cm]
        0&1&29.16\\[0.1cm]
        1&1&29.51\\[0.1cm]
        1.5&1&29.64\\[0.1cm]
        1&1.5&\textbf{29.87}\\[0.1cm]
        \bottomrule
    \end{tabular}
    \caption{The scores obtained with different values of $\lambda_1$ and $\lambda_2$. We train a Telugu Model and evaluate it on the telugu code-switched evaluation set.}
    \label{tab:lambda}
\end{table}

\section{Impact of Alignment Noise on \ourmodel's Performance}

In \ourmodel, we align speech representations with corresponding text embeddings. We use forced alignment to determine the number of speech embeddings associated with each text embedding. We introduce varying levels of noise into the alignment process during training and examine the effects on the model’s performance.
We begin with the current forced alignment and add noise to each alignment index $I$ using the formula $\lfloor I + N(0, \sigma) \rfloor$, where $N(0, \sigma)$ is a Gaussian distribution with mean 0 and standard deviation $\sigma$. Let us consider an example. Consider an original alignment of (2, 5, 8, 11), which indicates that for a given speech sequence $s_1$ to $s_{13}$ and a text sequence $w_1$ to $w_4$: $s_1$ to $s_2$ maps to $w_1$, $s_3$ to $s_5$ maps to $w_2$, $s_6$ to $s_8$ maps to $w_3$, and $s_9$ to $s_{11}$ maps to $w_4$. Now, adding noise to (2, 5, 8, 11) might yield (3, 6, 8, 12). Consequently, the new alignment would be: $s_1$ to $s_3$ maps to $w_1$, $s_4$ to $s_6$ maps to $w_2$, $s_7$ to $s_8$ maps to $w_3$, and $s_9$ to $s_{12}$ maps to $w_4$. If any text embeddings are leftover, we just use the last speech embedding for all the leftovers. Three different values of $\sigma$ ($\sigma = 1,3,5$) were tested to generate different levels of alignment noise.We conduct this experiment on Hindi Model trained with our 30 hr training data, and evaluate using the code-switched evaluation set.  We see in Table~\ref{tab:noisy_alignment} that the BLEU score degrades with the increase in $\sigma$ (increase in the noise). 

\begin{table}[h!]
    \small
    \centering
    \setlength\tabcolsep{3pt} 
    \begin{tabular}{lcc}
        \toprule
        \multicolumn{1}{c}{\textbf{$\sigma$}} & \multicolumn{1}{c}{\textbf{BLEU}} \\
        \midrule
        $\sigma=0$ & 33.12 \\
        $\sigma=1$ & 30.21 \\
        $\sigma=3$ & 27.68 \\
        $\sigma=5$ & 22.37 \\
        
        \bottomrule
    \end{tabular}
    \caption{BLEU scores of the \ourmodel model trained with different levels of alignment noise. The standard Hindi \ourmodel model with no added alignment noise (\(\sigma = 0\)) is compared against three models trained with varying degrees of noise (\(\sigma = 1\), \(\sigma = 3\), and \(\sigma = 5\)).}
    \label{tab:noisy_alignment}
\end{table}

\section{Dataset Annotation Guidelines}
\label{sec:annot}
\subsection{Code-switched and Monolingual Evaluation sets}

For both the code-switched and monolingual evaluation sets, approximately two hours of speech-transcription data were extracted for each of Telugu, Hindi, Marathi, and Bengali from IndicVoices \cite{javed2024indicvoices} for the code-switched evaluation set. Additionally, two hours of monolingual data were extracted specifically from IndicVoices for Telugu and Hindi for the Monolingual evaluation set. The transcripts were translated using IndicTrans2 \cite{gala2023indictrans}, with manual verification required to correct any errors in the machine-generated translations. 

The project cost for each language is as follows: Hindi - Rs. 3000 per hour, Marathi - Rs. 3000 per hour, Bengali - Rs. 3000 per hour, Telugu - Rs. 3500 per hour.

During the post-editing task, annotators who were the native speakers of the languages in the consideration were instructed to remove disfluencies and convert words entirely in uppercase to lowercase. The annotation process included: marking audio and transcription pairs as mismatches without editing if they were completely discordant, editing translations based on audio content in cases of minor mismatches between audio and transcription and excluding non-speech words from the translation process.

\subsection{Podcast Evaluation set}
For the podcast evaluation set, annotators were instructed to annotate transcripts of podcast speech and generate English translations after removing disfluencies from the corresponding transcripts.

\textbf{Guidelines for the tasks:}

The intended final dataset:
\begin{enumerate}
    \item Code-switched transcriptions with time markers, for Telugu and Hindi.
    \item Disfluency correction.
    \item English translations.
\end{enumerate}

\textbf{Guidelines for code-switched transcriptions:}
\begin{itemize}
    \item Maintain speaker turns to reflect continuous speech segments from individual speakers. Use speaker identifiers like "A" for the first speaker, "B" for the second, etc. Timestamps for these turns are necessary for aligning with audio clips.
    \item Transcribe disfluencies faithfully without correcting or omitting them.
    \item For intra-word code-switched words, retain the respective language script for each element.
    \item Indicate non-verbal sounds such as laughter using appropriate symbols or descriptors.
\end{itemize}

\textbf{Guidelines for disfluency correction:}
\begin{itemize}
    \item Correct only disfluent sentences; do not introduce additional words or change word order.
    \item Focus solely on removing disfluent words while preserving the original sentence's structure and meaning.
\end{itemize}

\textbf{Guidelines for English translation:}
\begin{itemize}
    \item Create a parallel dataset where each transcript is translated into fluent English, regardless of its disfluency status.
    \item Ensure translations accurately convey the original speech's meaning in natural, fluent English.
\end{itemize}

Detailed guidelines document can be found \href{https://docs.google.com/document/d/1179-T3nIUsQmybx2BvhNJIqx0kXx5ZtOxQqpoPqodhg/edit?usp=sharing}{here}. The transcription and translations required 4-5 rounds of verification. The cost for both the languages for all these tasks came out to be Rs.6000 per hour of audio.

\section{Experimental Details}
We fine-tune \ourmodel with a learning rate of $6e-5$. We use raw 16kHz speech as input to our model, and we jointly tokenize bilingual text using SentencePiece \cite{kudo-richardson-2018-sentencepiece}. We use Adam optimizer with parameters $\beta_1=0.9$, $\beta_2=0.98$, and a 20k-step warm-up period. A dropout rate of 0.15 is applied during training. We conducted experiments using Nvidia DGX A100 GPUs. We use sacreBLEU \cite{post-2018-call} to evaluate case-sensitive detokenized BLEU.


\end{document}